\documentclass[sigconf,nonacm]{acmart}

\usepackage{xcolor}

\AtBeginDocument{%
  \providecommand\BibTeX{{%
    \normalfont B\kern-0.5em{\scshape i\kern-0.25em b}\kern-0.8em\TeX}}}

\setcopyright{acmcopyright}
\copyrightyear{2023}
\acmYear{2023}
\acmDOI{XXXXXXX.XXXXXXX}

\acmConference[Multimodal KDD '23]{International Workshop on Multimodal Learning
}{August 07,
  2023}{Long Beach, CA}
\usepackage{caption}
\usepackage{subcaption}

\begin{document}

\title{Evaluating Picture Description Speech for Dementia Detection using Image-text Alignment}




\author{Youxiang Zhu}
\affiliation{%
  \institution{University of Massachusetts Boston}
  \city{Boston}
  \state{MA}
  \country{USA}
}
\email{Youxiang.Zhu001@umb.edu}

\author{Nana Lin}
\affiliation{%
  \institution{University of Massachusetts Boston}
  \city{Boston}
  \state{MA}
  \country{USA}
}
\email{Nana.Lin002@umb.edu}

\author{Xiaohui Liang}
\affiliation{%
  \institution{University of Massachusetts Boston}
  \city{Boston}
  \state{MA}
  \country{USA}
}
\email{Xiaohui.Liang@umb.edu}

\author{John A. Batsis}
\affiliation{%
  \institution{University of North Carolina}
  \city{Chapel Hill}
  \state{NC}
  \country{USA}
}
\email{John.Batsis@unc.edu}

\author{Robert M. Roth}
\affiliation{%
  \institution{Geisel School of Medicine at Dartmouth}
  \city{Lebanon}
  \state{NH}
  \country{USA}
}
\email{Robert.M.Roth@hitchcock.org}

\author{Brian MacWhinney}
\affiliation{%
  \institution{Carnegie Mellon University}
  \city{Pittsburgh}
  \state{PA}
  \country{USA}
}
\email{macw@andrew.cmu.edu}

\renewcommand{\shortauthors}{Youxiang Zhu et al.}

\begin{abstract}
  Using picture description speech for dementia detection has been studied for 30 years. Despite the long history, previous models focus on identifying the differences in speech patterns between healthy subjects and patients with dementia but do not utilize the picture information directly. In this paper, we propose the first dementia detection models that take both the picture and the description texts as inputs and incorporate knowledge from large pre-trained image-text alignment models. We observe the difference between dementia and healthy samples in terms of the text's relevance to the picture and the focused area of the picture. We thus consider such a difference could be used to enhance dementia detection accuracy. Specifically, we use the text's relevance to the picture to rank and filter the sentences of the samples. We also identified focused areas of the picture as topics and categorized the sentences according to the focused areas. We propose three advanced models that pre-processed the samples based on their relevance to the picture, sub-image, and focused areas. The evaluation results show that our advanced models, with knowledge of the picture and large image-text alignment models, achieve state-of-the-art performance with the best detection accuracy at 83.44\%, which is higher than the text-only baseline model at 79.91\%. Lastly, we visualize the sample and picture results to explain the advantages of our models.
\end{abstract}




\maketitle

\section{Introduction}
Dementia is a common and irreversible disease that affects more than 6 million older adults in the United States~\cite{alzheimer2022}. The speech-based analysis enables the detection of dementia in the early stage at a lower cost and lower effort compared to other alternative detection methods. Researchers have explored speech-based dementia detection via cookie theft picture description task for 30 years~\cite{becker1994natural}. In such a task, participants describe the cookie theft picture using spontaneous speech. The audio samples and the human-transcribed text samples are used to infer participants' cognitive health status, either Healthy Control (HC) or Alzheimer's Disease (AD). A significant challenge of dementia detection, different from speech and language research, is that labels are not a property of the data (i.e., speech) and can only be obtained from the participants at the moment. Researchers aim to develop models to infer cognitive labels from audio and text samples. Specifically, researchers explored handcrafted and automatic acoustic and linguistic features and discovered that the linguistic features of text samples produced the most effective dementia detection models~\cite{cummins2020comparison, koo2020exploiting, edwards2020multiscale, zhu2021exploring}. 

\begin{figure}
  \centering
  \includegraphics[width=0.45\textwidth]{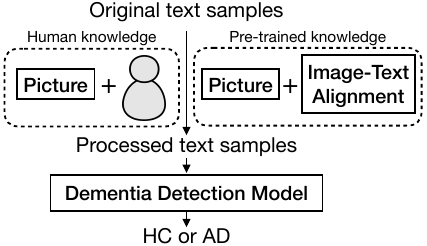}
  \caption{Human knowledge vs. Pre-trained knowledge}
  \label{fig:overview}
  \vspace{-0.5cm}
\end{figure}


{Picture information has been incorporated into the model in limited ways, e.g., information units~\cite{yancheva-rudzicz-2016-vector}, dialogue acts~\cite{farzana-etal-2020-modeling} and eye tracking~\cite{barral2020non}, as shown on the left of Figure~\ref{fig:overview}. These models explore the knowledge of the picture extracted by humans and have not obtained full knowledge of the picture. The information units are either defined as a set of words or phrases manually or automatically using participants' text samples. The dialogue acts are generated based on fixed areas of the picture and used to label the sentences with human effort. The eye tracking methods represent the gaze features by manually defining 13 areas of interest on the image while having no knowledge about the image contents in the areas. The lack of the picture as an input to the model prevents the model from accessing the full and original knowledge of the picture. While deep learning models rapidly advance beyond human capability, we envision that dementia detection models using the original picture as an input can understand the picture description task deeply and, as a result, outperform previous models that do not use the picture as a direct input.}

The image-text alignment models are advanced recently~\cite{radford2021learning, li2022grounded, zhang2022glipv2, wehrmann2020adaptive} and have been successfully applied to many domains, e.g., image-text retrieval~\cite{sun-etal-2021-lightningdot} and multi-modal sarcasm detection~\cite{liu2022towards}. It can evaluate the relevance between a set of images and a set of texts. With such models available, our preliminary results indicate the difference between AD and HC samples in terms of the text’s relevance to the picture and the focused area of the picture. We envision utilizing such differences could help enhance the detection accuracy. To this end, we propose advanced models that pre-process the samples based on the image-text alignment information.

Our contributions are three-fold.

First, we analyze the different relevance of HC and AD samples to the picture using image-text alignment and have two observations: HC participants speak less quantity but more quality samples, and in the description processes, HC participants focus on two more areas of the picture than AD, i.e., the faucet area and the area outside of the window.

Second, we propose three advanced models, i) using the picture relevance to filter sentences of samples, ii) using the dementia-sensitive sub-image to filter sentences of samples, and iii) using the most text-relevant focused areas as topics to organize the sentences of samples. While the baseline model takes samples as inputs, our three advanced models take the samples processed using the picture and image-to-text alignment models as inputs.

Third, we conduct extensive experiments to evaluate the proposed three advanced models. The results show that they have successfully explored the picture information and image-to-text alignment models to improve the accuracy from baseline 79.91\% to (80.63\% picture relevance, 83.44\% sub-image relevance, and 82.49\% focused area). We further show our models achieve higher or equal accuracy than existing works.

\section{Related work}

{\textbf{Speech and text learning for dementia detection.}}
Researchers exploited various speech tasks such as grocery shopping dialog~\cite{gong2022hong}, speech and writing~\cite{gkoumas2021longitudinal}, telephone interview~\cite{Brandt1988TheTI, konagaya2007validation} and voice assistants~\cite{liang2022evaluating, mirheidari2017avatar}. The picture description task using the cookie theft picture~\cite{becker1994natural, luz2020alzheimer, luz2021detecting} is one of the most popular speech tasks in dementia detection. Although such a task has been studied for 30 years, it suffers from limited data problems due to the high cost of data collection. To enable effective learning with small data, researchers applied deep transfer learning techniques and showed that automatic features are more effective than handcrafted features ~\cite{zhu2021exploring,balagopalan2020bert}. Recently, researchers further explored some more specific research directions to improve learning with small data, such as automatic speech recognition~\cite{wang2022conformer, vsvec2022evaluation}, data augmentation~\cite{bertini2021automatic, hledikova2022data}, intermediate pre-training~\cite{zhu2022domain}, incorporating pause information~\cite{farzana-etal-2022-say, yuan2021pauses} and 
prompt learning~\cite{wang2022exploiting}. Different from these directions, our work is the first to explore the picture information and knowledge from language and image-text alignment models for dementia detection. 

{\textbf{Picture information for dementia detection.} Previous works have explored “information units” from the cookie theft picture as a manually crafted feature to implement the classification. The information units are defined as a set of words or phrases either manually or automatically extracted from participants’ description texts. Importantly, Croisile et al.~\cite{croisile1996comparative} showed on average, AD participants produce 9.23 information units, while HC participants produce 14.46 information units. The difference is statistically significant. This evidence confirms that the relevance of HC descriptions to the picture should be higher than AD descriptions. 
Other methods like eye tracking and dialogue acts explore the visually or verbally focused areas in the picture. The dialogue acts~\cite{farzana-etal-2020-modeling} are generated based on eight areas of the picture and used to label the sentences with human effort. 
Eye-tracking~\cite{barral2020non} is based on 13 human-defined areas of interest to represent the eye-tracking features and then combine them with language features for classification purposes, employing early/late fusion techniques. We realize that using human efforts or adding another modality like eye-tracking comes with a high cost. In addition, human-defined, fixed areas are sub-optimal, which have limited consideration of the boundary of the objects in the picture, and the corresponding models have a limited understanding of the content in the areas.}

{Compared to the above methods, ours have the following advantages: i) Our model does not need any human efforts in feature engineering and labeling. Information units and dialogue acts need to be defined and labeled by humans, while our methods rely on the image-text alignment model, which can be done automatically. ii) Our model is more capable of processing picture information. Information unit using words or phrases to represent the objects in the picture. Dialogue act and eye-tracking used human-defined, fixed areas. In comparison, with the image-text alignment technique, our model processes all sub-images that may contain any objects and can analyze the details of the picture information automatically.}

{\textbf{Image-text alignment.} There are many pre-trained multi-modal models that emerged in the language model research area, e.g., CLIP~\cite{radford2021learning}, BLIP-2~\cite{li2023blip}, KOSMOS~\cite{huang2023language} and PaLM-E~\cite{driess2023palm}. They bridge the gap between images and text, allowing the model to comprehend and reason about visual content based on textual descriptions. This opens up possibilities for various applications that require understanding multimodal data, such as image captioning, visual question answering, and cross-modal retrieval. }

{Image-text alignment can be applied to multiple multi-modal tasks, such as image-text retrieval~\cite{sun-etal-2021-lightningdot} and multi-modal sarcasm detection~\cite{liu2022towards}.
One advantage of the image-text alignment models is their great zero-shot performance~\cite{zhou2022learning}. In other words, it can be used without further fine-tuning on the downstream tasks. We plan to apply this zero-shot advantage in dementia detection. Specifically, we envision the image-text alignment models can well understand the contents of the cookie theft picture and the description texts using the knowledge from large pre-trained datasets and produce accurate relevance between the picture and the description texts.}

\section{Background and preliminary study}

{We present the problem formation of dementia detection using the picture description dataset, definitions including relevance between images and texts, quantity and quality of samples, and focused areas of the picture, and preliminary results of our approaches.}

{\subsection{Problem formation}}
A picture description dataset for dementia detection $S_{x,y} =\\ \{(x_1, y_1), (x_2, y_2), \dots, (x_n, y_n)\}$ has $n$ pairs of human transcribed text samples and labels, where $x_i$ is a human-transcribed text sample of the cookie theft picture $g$ by the $i$-th participant and $y_i$ is the $i$-th participant's label, either HC or AD. We denote $S_x = \{x_1, x_2, \cdots, x_n\}$, $S_{x,HC}$ as a subset of samples of the HC, and $S_{x,AD}$ as a subset of samples of the AD. We have $S_x = S_{x,HC}\cup S_{x,AD}$. The dementia detection problem is to infer a label $y_i$ from a sample $x_i$.

\subsection{Relevance between images and texts}

CLIP model is a recent advance in multi-modal learning that can be used to measure the relevance between the images and texts~\cite{radford2021learning}. Formally, given $m$ images and $n$ texts as input, the CLIP model outputs a $m*n$ matrix $M$ that represents the relevance scores between the images and texts. Using this matrix, we define two methods to explore the relevance between images and texts. An \textbf{image-to-texts match} method aims to generate the relevance scores of one image and multiple texts. We use the vector of the $i$-th row in $M$ to derive the relevance between the $i$-th image and all texts. Specifically, a softmax function is used to convert the values in the vector to probabilities, and the probabilities are used as relevance scores. A \textbf{text-to-images match} method aims to generate the relevance scores of one text and multiple images. We use the vector of the $j$-th column in $M$ to derive the relevance between the $j$-th text and all images. The image-to-texts and text-to-image match methods can be used to find the relevant texts and images.

\subsection{Quantity and quality of samples}

\begin{table}[]
\centering
\resizebox{0.45\textwidth}{!}{%
\begin{tabular}{cccc}
\hline
   & Relevance & sentence num/sample & word num/sample \\ \hline
HC & $c_{HC} = 19.66$      & 16.52               & 144.28          \\
AD & $c_{AD}=14.57$      & 17.70               & 158.35          \\ \hline
\end{tabular}%
}
\caption{Preliminary results. Relevance scores are scaled by the total number of sentences in all samples.}
\label{tab:text-image-relevance}
\end{table}

In this section, we explore the relevance of the description samples of the HC and AD to the original cookie theft picture and investigate whether the relevance of the HC and AD show difference. {We aim to study two aspects: quantity and quality. The quantity is the number of word or sentences produced by participants. The quality is the relevance of the speech to the cookie theft picture.}

We define two relevance scores $(c_{HC}, c_{AD})$ to represent the relevance of all HC samples $S_{x,HC}$ and all AD samples $S_{x,AD}$ to the picture, respectively. We apply the \textbf{image-to-texts match} method to calculate the relevance score $c_{i,j}$ between the original cookie theft picture and a sentence $x_{i,j}$ of a sample $x_i$. The relevance score between a sample $x_i$ and the picture is then calculated as $c_i = \sum_{x_{i,j} \in x_i}{c_{i,j}}$. For all HC samples, we calculate the mean value $c_{HC}$ of all relevance scores $\{c_i | x_i\in S_{x, HC} \}$. For all AD samples, we calculate the mean value $c_{AD}$ of all relevance scores $\{c_i | x_i\in S_{x, AD} \}$ (shown in Table~\ref{tab:text-image-relevance}). We have two observations: 
i) $c_{HC} > c_{AD}$. ii) The numbers of sentences and words per sample in $S_{x,HC}$ are smaller than $S_{x,AD}$ . We conclude that, in general, HC participants produce lower quantity but higher quality samples than AD participants.

\subsection{Focused areas of picture}

\begin{figure*}
\begin{center}
  \begin{subfigure}{0.45\textwidth}
    \centering
    \includegraphics[width=\linewidth]{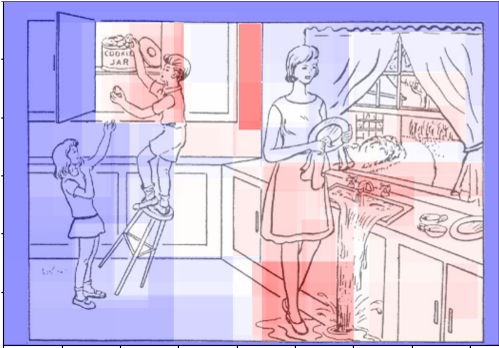}
  \end{subfigure}%
  \ \ \ 
  \begin{subfigure}{0.45\textwidth}
    \centering
    \includegraphics[width=\linewidth]{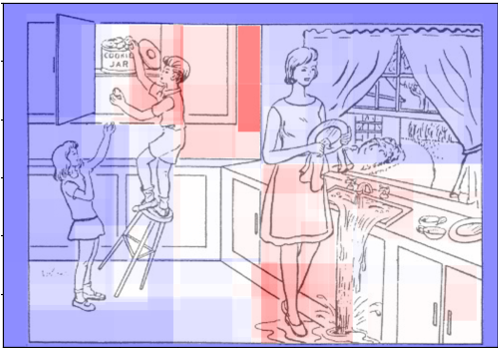}
  \end{subfigure}
  \caption{The focused area of HC (left) and AD (right). Red means highly focused and blue means lowly focused.}
  \label{fig:focus}
\end{center}
\end{figure*}

{“Focused areas” are the areas in the cookie theft picture that participants’ description texts are most relevant to. The focused areas in our paper are noticed and described by the participants. This is different from the visually focused areas where an eye-tracking device~\cite{barral2020non} is required to collect such information. Technically, we use the image-text alignment to identify the focused areas in the picture that have the highest relevance scores with all description texts.}
We aim to find the different focused areas of the picture between HC and AD participants. Specifically, we adopt the selective search method~\cite{UijlingsIJCV2013} to generate sub-images from the picture. Selective search has been commonly used for region proposals in object detection. For a sentence $x_{i,j}$ in a sample $x_i$, we use the \textbf{text-to-images match} method to find the sub-image that is the most relevant to $x_{i,j}$. We then merge the sub-images most relevant to the sentences of all samples in $S_{x,HC}$ in a heatmap, and merge the sub-images most relevant to the sentences of all samples in $S_{x,AD}$ in another heatmap, shown in Figure~\ref{fig:focus}. We have two observations: i) The common focused areas of HC and AD participants are cookie jar and water on the floor. ii) HC focuses on more areas than AD, i.e., the faucet area and the area outside of the window. 

Our preliminary results have shown both the image-to-texts and text-to-images match methods can reveal the different relevance of HC and AD samples to the picture, which may be further used to enhance dementia detection accuracy.

\section{Method}
We first propose a baseline {text-only} dementia detection model and then develop three advanced models using the 
relevance between images and texts.

\subsection{Baseline model}

Given a picture description dataset $S_{x,y}$, a baseline dementia detection model can be implemented in the following steps: i) for each sample $x_i$, we use a pre-trained language model (e.g., BERT~\cite{devlin2019bert}) to generate tokens of $x_i$ and generate embedding of each token. An embedding $e_i$ of $x_i$ is defined as the average embedding of all tokens of $x_i$; ii) we input embedding $e_i$ and label $y_i$ (either HC or AD) to develop a classification model, e.g., SVM. The baseline model will be used as a baseline for performance comparison and used as a component of the advanced models where our text-to-images and image-to-texts match methods will process the samples to improve the performance.

\subsection{Picture relevance model}

The picture description samples from HC and AD may have raw and noisy segments that positively or negatively impact dementia detection. As we have successfully shown that the relevance scores between the picture and the samples from HC and AD are different, we aim to investigate the following research question: can we use the picture to filter segments of samples to enhance accuracy?

To this end, we apply the \textbf{image-to-texts match} method using original cookie theft picture $g$ and each sentence $x_{i,j}$ of a sample $x_i$ to generate a relevance score $c_{i,j}$. By sorting $c_{i,j}$, we find out the top-$k_t$ and bottom-$k_b$ sentences of the sample $x_i$ relevant to the picture $g$. We denote the selection process of the top-$k_t$ and bottom-$k_b$ sentence as $\delta$, and we have $\bar{x}_i = \delta_{(k_t, k_b)}(x_i, g)$ to represent a subset of sentences most relevant and most irrelevant to the picture. We then concatenate the sentences using their original order. If the total number of sentences in a sample is smaller than $k_t + k_b$, we filter out the non-top-$k_t$ and non-bottom-$k_b$ sentences and ensure each sentence appears only once in the processed sample. Finally, the processed samples and the corresponding labels are used to develop the baseline dementia detection model.

\subsection{Sub-image relevance model}

The cookie theft picture has many objects. Description texts relevant to sub-images with different objects may result in different dementia detection performances. Thus, we aim to find a \textit{dementia-sensitive} sub-image where the sentences filtered using their relevance to this sub-image might result in enhanced dementia detection accuracy. Specifically, inspired by the R-CNN object detection pipeline~\cite{girshick2014rich}, we first generate a set of sub-images using the selective search~\cite{UijlingsIJCV2013}. These sub-images are expected to be high recall for finding objects. For each sub-image $g_s$ and samples $x_i\in S_x$, we derive $\bar{S}_{x,s} = \{\bar{x}_i = \delta_{(k_t, k_b)}(x_i, g_s)$ for $x_i\in S_x\}$, where the processed sample $\bar{x}_i$ includes $k_t$ most relevant and $k_b$ most irrelevant sentences of all samples to the sub-image $g_s$. Then, we extract the embedding $e_{i,s}$ for each processed sample $\bar{x}_{i}$ (with label $y_i$), and calculate the pair-wise cosine similarity as
\[
    d_s = \sum_{y_i=y_{i'}}{cos(e_{i,s}, e_{i',s})} - \sum_{y_i \neq y_{i'}}{cos(e_{i,s}, e_{i',s})}
\]
We aim to maximize $d_s$ by maximizing the cosine similarity of embedding of the same label and minimizing the cosine similarity of embedding of different labels. After calculating $d_s$ for all sub-images, we define the dementia-sensitive sub-image $g_s$ as the sub-image with the maximum score $d_s$, derive $\bar{S}_{x,s} = \{\bar{x}_i = \delta_{(k_t, k_b)}(x_i, g_s)$ for $x_i \in S_x\}$, extract the embedding $e_{i,s}$ of the processed samples $\bar{x}_i$, and use embedding $e_{i,s}$ and the label $y_i$ to develop the baseline dementia detection model.

\subsection{Focused area model}

Previous works~\cite{yancheva-rudzicz-2016-vector, croisile1996comparative, lai2009semantically} extracted information units from the samples and used information units as topics. We are the first to explore the topics using the focused areas of the cookie theft picture, each focused area corresponding to a topic. Specifically, we use selective search to generate sub-images. For each sentence of all samples in $S_x$, we calculate its relevance score with every sub-image $g_s$ using the \textbf{text-to-images match} method. Then we sum up the relevance scores according to each sub-image. To select the focused areas from these sub-images, we first perform non-maximum suppression to filter out similar sub-images that have lower summed scores, and then we select top-$k_f$ sub-images with the highest summed relevance scores as focused areas. The $k_f$ focused areas are denoted as $G = \{g_1, g_2, \dots g_{k_f}\}$. We treat each focused area as a topic and organize the sentences using these topics. For each sentence in a sample $x_i$, we match it to one focused area in $G$ that has the highest relevance score using the \textbf{text-to-images match} method. In other words, we organize the sentences in $x_i$ into $k_f$ categories. For each category, we concatenate the corresponding sentences and obtain their embedding. Finally, we concatenate the embedding of all the topics and used the concatenated embedding to develop the baseline dementia detection model. 

\begin{figure}
  \centering
  \includegraphics[width=0.45\textwidth]{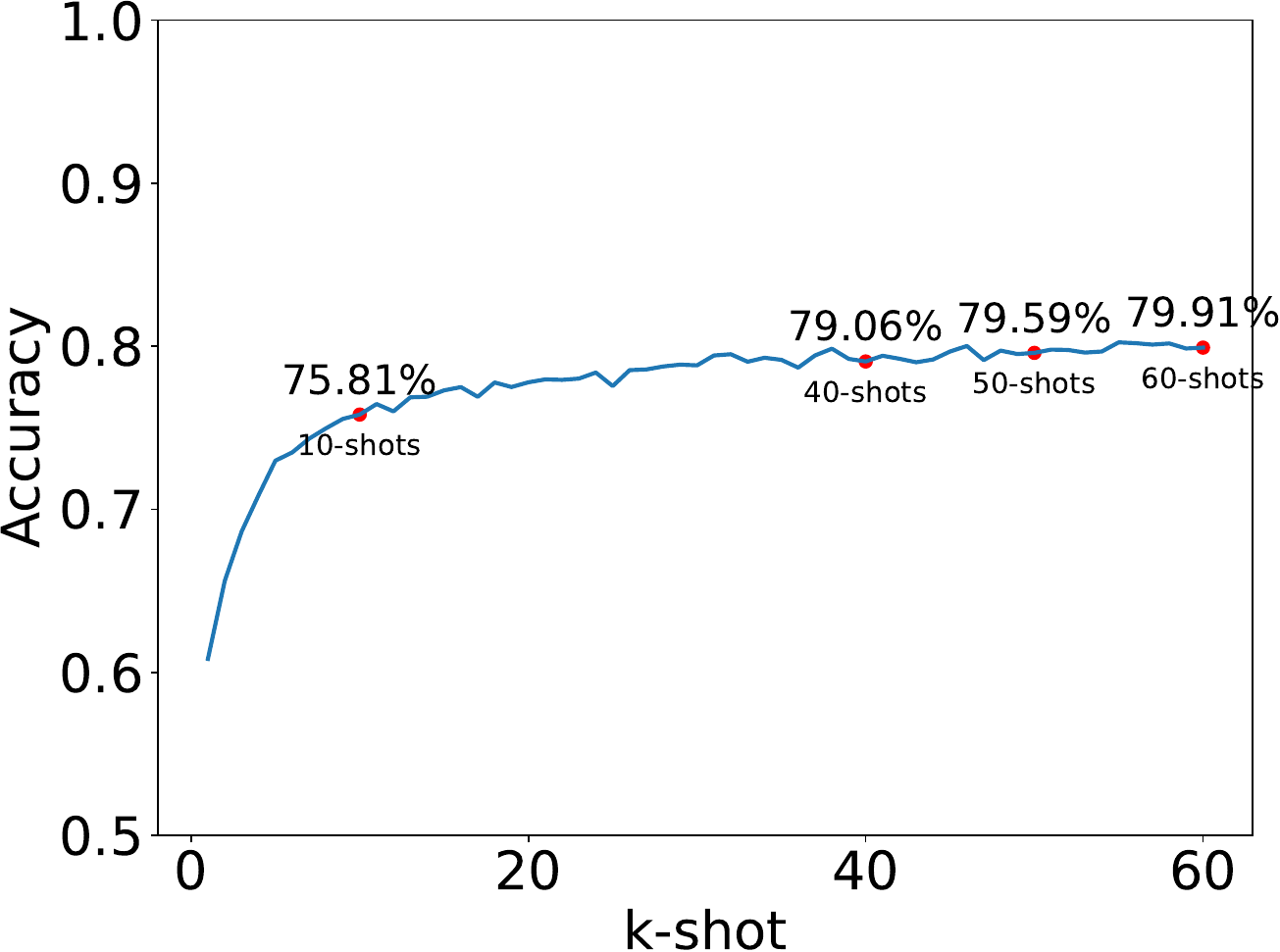}
  \caption{The accuracy result of baseline model}
  \label{fig:baseline}
\end{figure}

\begin{figure*}
  \begin{subfigure}{0.49\textwidth}
    \centering
    \includegraphics[width=\linewidth]{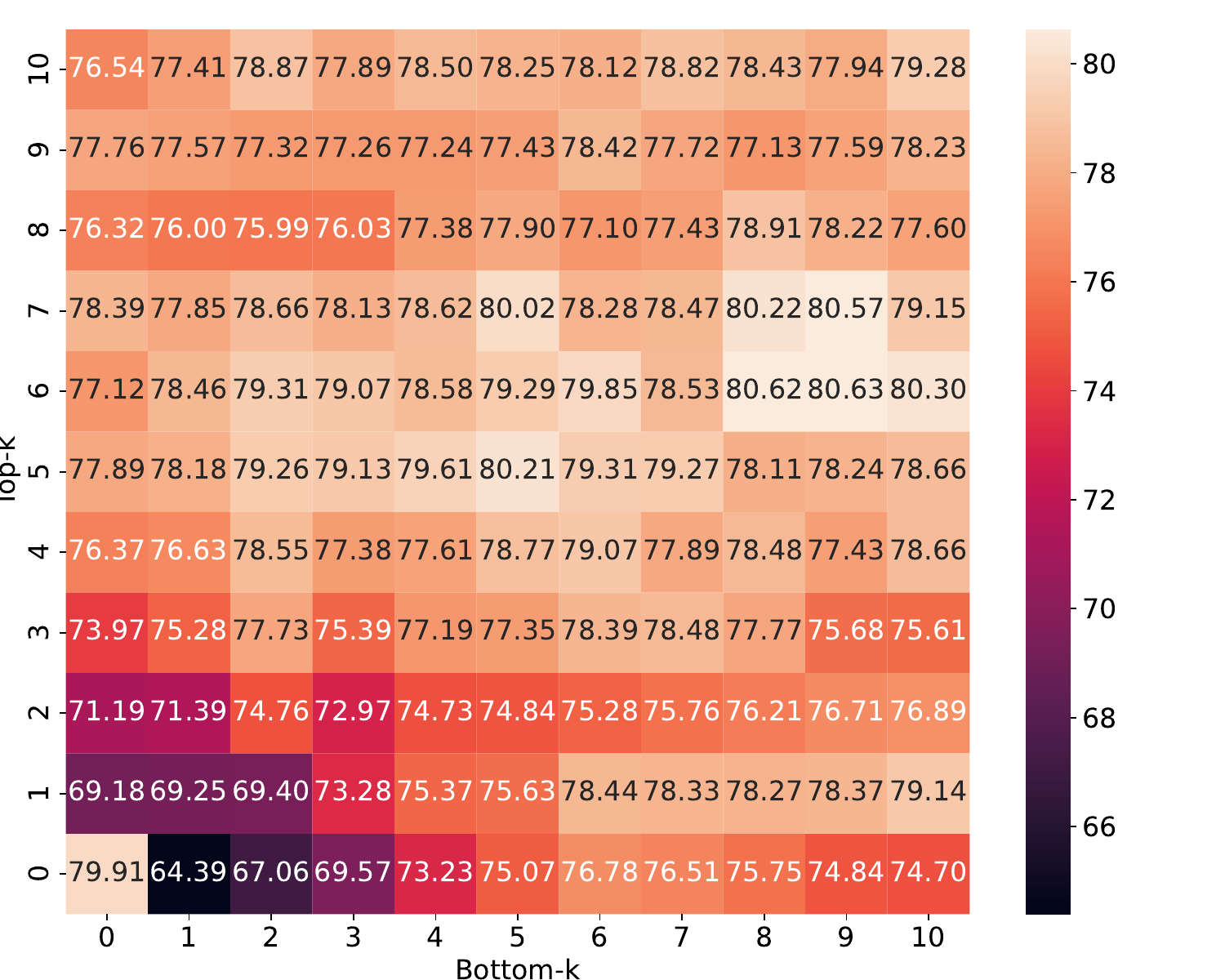}
    \caption{Picture relevance model}
    \label{fig:adv_picture}
  \end{subfigure}%
  \begin{subfigure}{0.49\textwidth}
    \centering
    \includegraphics[width=\linewidth]{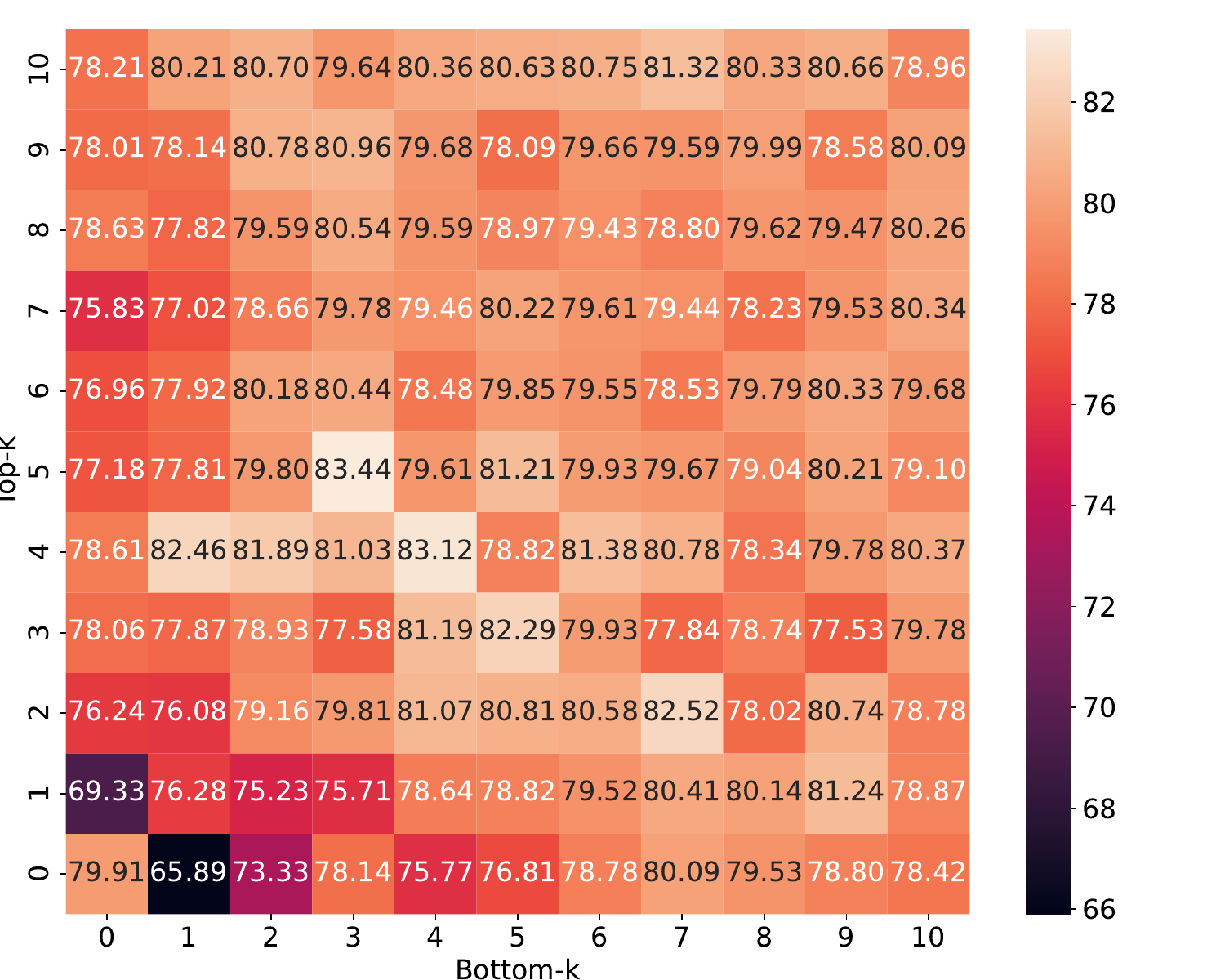}
    \caption{Sub-image relevance model}
    \label{fig:adv_sub_image}
  \end{subfigure}
  \caption{Results of 60-shots evaluation}
  \label{fig:focus}
\end{figure*}

\section{Experiments}

We introduce the experimental data, implementation details, evaluation protocol, and evaluation results of our models.

\subsection{Data}

ADReSS ~\cite{luz2020alzheimer} is a cookie theft picture description dataset in English. It was processed based on the Pitt Corpus dataset~\cite{becker1994natural} with the balanced label, age, and gender. All samples are human-transcribed description texts. Each sample has a label, either HC or AD. ADReSS has 108 samples for training and 48 samples for testing. 

\subsection{Implementation detail}
{
We used publicly available codes of BERT\footnote{\url{https://huggingface.co/bert-base-uncased}}, CLIP\footnote{\url{https://huggingface.co/openai/clip-vit-base-patch32}} and scikit-learn\footnote{\url{https://scikit-learn.org/}} to implement our dementia detection models. We use the default hyperparameters for the baseline model. We run all the experiments with a single V100 GPU. }

\subsection{Few-shot evaluation protocol}

Inspired by the evaluation protocol used in few-shot learning~\cite{guo2020broader}, we propose an evaluation protocol for dementia detection with limited data. We first combine the original training and testing datasets into one dataset (e.g., 156 samples in ADReSS). We consider the $2$-way $k$-shot setting for our binary classification task. For each round, we randomly select $k$ samples of each class for training and randomly select another 15 different samples of each class for testing. We repeat this process for 600 rounds and report the average performance. 

We consider other evaluation protocols, e.g., cross-validation or fixed training testing split have disadvantages: cross-validation may overestimate the performance~\cite{tsamardinos2015performance}, and fixed training testing splits are also infeasible for such a small dataset. In ADReSS 2020~\cite{luz2020alzheimer}, the standard evaluation protocol uses 48 samples for testing. The different results on one sample can lead to around $2\%$ accuracy difference. Thus, such evaluation produces unstable results. Our proposed evaluation protocol achieves 79.67\% accuracy in a 54-shots setting, while in the original ADReSS evaluation protocol, the number of training samples for each class is 54 and the accuracy on the testing set is 83.33\%. We consider the original ADReSS evaluation protocol overestimates the performance, and our proposed evaluation protocol provides a more representative result.

\subsection{Results of baseline model}

We report the 1-60 shots results of the baseline model on ADReSS dataset in Figure \ref{fig:baseline}. We observe that the accuracy increases rapidly from 1-10 shots (60.82\% to 75.81\%), the accuracy increases slowly from 10-40 shots (75.81\% to 79.06\%), and the accuracy saturates after 40-shots. The accuracy of 40-shots, 50-shots, 55-shots and 60-shots are 79.06\%, 79.58\%, 80.23\%, and 79.91\%, respectively. The results suggested that adding more samples after 40-shots in training may lead to limited accuracy improvement.

\subsection{Results of picture relevance model}

We evaluate our picture relevance model with parameters $k_t = [0,10],k_b=[0,10]$ using the 60-shots evaluation protocol. The accuracy results are shown in Figure~\ref{fig:adv_picture} where the position $(0, 0)$ is the result of the baseline model using all the sentences. We observe that:
i) The best accuracy is achieved with $(k_t,k_b)=(6,9)$ (80.63\%) statistical significant with t-test p=0.008 < 0.01 compared to the baseline model. 
ii) Using only bottom-$k_b$ sentences but not using top-$k_t$ sentences resulted in the best accuracy of 76.78\%, worse than the baseline 79.91\%, which implies the highly picture-relevant sentences play an important role in dementia detection.
iii) Using only top-$k_t$ sentences but not using bottom-$k_b$ sentences resulted in the best accuracy at 78.39\% slightly worse performance than the baseline 79.91\%, which implies the effectiveness of picture-irrelevant sentences in dementia detection.
iv) Using top-$k_t$ ($5 \leq k_t \leq 7$) and bottom-$k_b$ ($k_b \geq 5$) resulted in equal or higher accuracy than the baseline model, which confirms the effectiveness of our proposed filtering process based on picture relevance.

\textbf{Sample Visualization.} We show the details of the processed samples in Table~\ref{table:samples}. The picture-irrelevant sentences include other dialog acts such as acknowledgment, instruction, question and answering, stalling, and so on~\cite{farzana-etal-2020-modeling}. For example, the research assistant may say, "just tell me all of the action" and "okay good". And the participants may say "and that’s it." Such non-picture-description dialog acts are irrelevant to the picture, but could still be effective in dementia detection. By looking at the samples, we found that AD participants spoke more picture-irrelevant sentences than HC participants, and our advanced model took advantage of these sentences.

\begin{table*}
\begin{center}
\begin{small}
\resizebox{\textwidth}{!}{ 
\begin{tabular}{ p{0.04\textwidth}p{0.32\textwidth}p{0.32\textwidth}p{0.32\textwidth}} 
 \hline
 ID & Processed samples of the picture relevance model. \textcolor{red}{Red}: top-5 sentences. \textcolor{blue}{Blue}: bottom-5 sentences. & Processed samples of the sub-image relevance model. \textcolor{red}{Red}: top-5 sentences. \textcolor{blue}{Blue}: bottom-3 sentences. & Processed samples of focused area model. \textcolor{red}{Red}: focused area 1. \textcolor{blue}{Blue}: focused area 3.\\ 
 \hline
 S207 (HC) & \textcolor{blue}{ just tell me all of the action}. \textcolor{blue}{ little girl with her finger to her lips}. \textcolor{red}{ the boy on the stool}. \textcolor{red}{ stool tipping over}. \textcolor{red}{ getting cookies out of the cookie jar}. \textcolor{red}{ uh mother washing dishes}.  water running. \textcolor{red}{ sink overflowing}. \textcolor{blue}{ xxx those curtains are blowing or not}. \textcolor{blue}{ that's about it}. \textcolor{blue}{ okay good}. &  \textcolor{red}{ just tell me all of the action}.  little girl with her finger to her lips. \textcolor{red}{ the boy on the stool}. \textcolor{red}{ stool tipping over}. \textcolor{red}{ getting cookies out of the cookie jar}. \textcolor{red}{ uh mother washing dishes}. \textcolor{blue}{ water running}. \textcolor{blue}{ sink overflowing}. \textcolor{blue}{ xxx those curtains are blowing or not}.  that's about it.  okay good. & \textcolor{red}{ just tell me all of the action}. \textcolor{red}{ little girl with her finger to her lips}. \textcolor{blue}{ the boy on the stool}. \textcolor{red}{ stool tipping over}. \textcolor{blue}{ getting cookies out of the cookie jar}. \textcolor{red}{ uh mother washing dishes}. \textcolor{red}{ water running}. \textcolor{red}{ sink overflowing}. \textcolor{red}{ xxx those curtains are blowing or not}. \textcolor{red}{ that's about it}. \textcolor{red}{ okay good}.  \\ 
 \hline
 S162 (AD) & in the picture. \textcolor{red}{ I see uh two kids up at the cookie jar, one on a stool the other standing on the floor}.  cupboard door is opened. \textcolor{red}{ mother's washing the dishes}.  the water is running overflowing the sink. \textcolor{red}{ and uh there's two cups and a plate on the counter}. \textcolor{red}{ and she's washing holding a plate in her hand}. \textcolor{blue}{ curtains at the windows}.  the cookie jar has the lid off. \textcolor{blue}{ hm hm that's about it}.  cupboards underneath the sink.  cupboards underneath the other cupboards. \textcolor{red}{ uh kid falling off the stool}.  the girl laughing at him.  cookies in the cookie jar with the lid off. \textcolor{blue}{ he has a cookie in his hand}. \textcolor{blue}{ and that's it}. \textcolor{blue}{ okay good}. & in the picture. \textcolor{red}{ I see uh two kids up at the cookie jar, one on a stool the other standing on the floor}. \textcolor{red}{ cupboard door is opened}. \textcolor{red}{ mother's washing the dishes}. \textcolor{blue}{ the water is running overflowing the sink}.  and uh there's two cups and a plate on the counter.  and she's washing holding a plate in her hand. \textcolor{blue}{ curtains at the windows}.  the cookie jar has the lid off.  hm hm that's about it.  cupboards underneath the sink.  cupboards underneath the other cupboards. \textcolor{red}{ uh kid falling off the stool}. \textcolor{red}{ the girl laughing at him}.  cookies in the cookie jar with the lid off. \textcolor{blue}{ he has a cookie in his hand}.  and that's it.  okay good. & \textcolor{red}{ in the picture}. \textcolor{blue}{ I see uh two kids up at the cookie jar, one on a stool the other standing on the floor}. \textcolor{blue}{ cupboard door is opened}. \textcolor{red}{ mother's washing the dishes}. \textcolor{red}{ the water is running overflowing the sink}. \textcolor{blue}{ and uh there's two cups and a plate on the counter}. \textcolor{red}{ and she's washing holding a plate in her hand}. \textcolor{blue}{ curtains at the windows}. \textcolor{blue}{ the cookie jar has the lid off}. \textcolor{red}{ hm hm that's about it}. \textcolor{red}{ cupboards underneath the sink}. \textcolor{blue}{ cupboards underneath the other cupboards}. \textcolor{red}{ uh kid falling off the stool}. \textcolor{red}{ the girl laughing at him}. \textcolor{blue}{ cookies in the cookie jar with the lid off}. \textcolor{blue}{ he has a cookie in his hand}. \textcolor{red}{ and that's it}. \textcolor{red}{ okay good}.    \\ 
 
 \hline
\end{tabular}
}
\caption{Sample visualization}
\label{table:samples}
\end{small}
\end{center}
\end{table*}

\subsection{Results of sub-image relevance model}

\begin{table*}[]
\centering
\resizebox{\textwidth}{!}{%
\begin{tabular}{ccccccccc}
\hline
Top-k-bottom-k & 1 & 5 & 10 & 20 & 30 & 40 & 50 & 60 \\ \hline
 Baseline & ${60.82}_{12.10}$ & ${72.98}_{8.71}$ & ${75.81}_{7.63}$ & ${77.78}_{7.14}$ & ${78.82}_{6.92}$ & ${79.06}_{6.32}$ & ${79.59}_{6.47}$ & ${79.91}_{7.05}$ \\
(6, 9)-picture & ${60.76}_{10.99}$ & ${72.46}_{8.36}$ & ${75.39}_{7.49}$ & ${77.89}_{7.33}$ & ${79.38}_{6.71}$ & ${79.38}_{7.18}$ & ${80.38}_{6.80}$ & ${80.63}_{6.56}$ \\ 
(5, 3)-sub-image & ${63.08}_{12.66}$ & ${75.07}_{8.46}$ & ${78.86}_{6.86}$ & ${81.37}_{6.51}$ & ${81.64}_{6.38}$ & ${82.22}_{6.06}$ & ${82.98}_{6.26}$ & ${83.44}_{6.36}$ \\ \hline

\end{tabular}%
}
\caption{Comparison between baseline model, picture relevance model, and sub-image relevance model}
\label{tab:adv_sub_image}
\end{table*}

Similarly, we evaluate the model with parameters $k_t = [0,10]$ and $k_b=[0,10]$ using 60-shots evaluation protocol. For each case, we report the result of the sub-image with the highest score $d_s$ in Figure~\ref{fig:adv_sub_image}. We observed that the sub-image relevance model requires less number of sentences to achieve higher accuracy than the picture relevance model.  The sub-image relevance model achieved the highest accuracy 83.44\% with $(k_t,k_b)=(5,3)$, while the picture relevance model achieved the highest accuracy at 80.63\% with $(k_t,k_b)=(6,9)$. It confirms that the relevance to the dementia-sensitive sub-image is a more effective metric than the relevance to the entire picture for dementia detection. Also, as shown in Table \ref{tab:adv_sub_image}, the sub-image model requires fewer shots to the same accuracy compared to the baseline and picture’s relevance model. 

\begin{figure}
  \begin{subfigure}{0.25\textwidth}
    \centering
    \includegraphics[width=\linewidth]{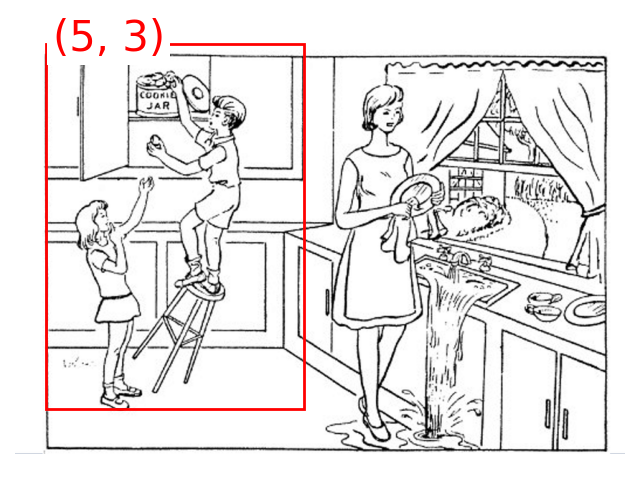}
    \caption{Sub-image of $(5, 3)$}
    \label{fig:adv_sub_image_vis}
  \end{subfigure}%
  \begin{subfigure}{0.25\textwidth}
    \centering
    \includegraphics[width=\linewidth]{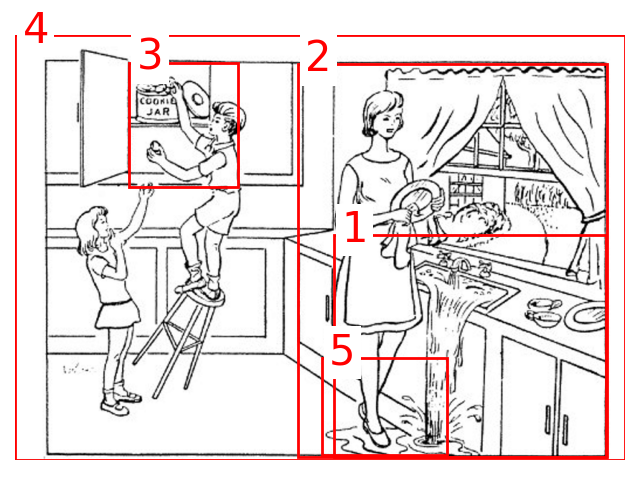}
    \caption{Top-5 focused areas}
    \label{fig:topic}
  \end{subfigure}
  \caption{Picture visualization}
  
  \label{fig:visualization}
\end{figure}

\textbf{Picture visualization.} In the best accuracy case (83.44\% with $(k_t,k_b)=(5,3)$), we found that the dementia-sensitive sub-image located on the left part of the picture, as shown in Figure~\ref{fig:adv_sub_image_vis}. In addition, other results close to the best accuracy use $(4, 2)$, $(4, 4)$, $(4, 6)$, $(4, 7)$, use this same sub-image, which reveals that the left part of the picture is the most dementia-sensitive.

\textbf{Sample visualization.} Table~\ref{table:samples} shows the processed samples. As we use the dementia-sensitive sub-image (left part of the picture), compared to the processed sample using the picture relevance, the sentences "and that’s it" and "okay good" no longer appear in the bottom-$k_b$ sentences; instead, the sentences relevant to the right part of the picture are considered as the bottom-$k_b$ sentences, e.g., "the water is running overflowing the sink" and "water running". Note that, our model takes both top-$k_t$ and bottom-$k_b$ sentences as inputs, and using the sub-image relevance may improve the quality of the processed samples and enhance the accuracy.

\subsection{Results of focused area model}

\begin{table}[]
\centering
\resizebox{0.5\textwidth}{!}{%
\begin{tabular}{cccc}
\hline
Areas & 60-shots accuracy & \multicolumn{1}{l}{Areas} & \multicolumn{1}{l}{60-shots accuracy} \\ \hline
Baseline & ${79.91}_{7.05}$ & &\\
\hline
(1, 2) & ${80.83}_{6.62}$ & (1, 2, 3) & ${82.24}_{5.93}$ \\
(1, 3) & ${82.49}_{6.34}$ & (1, 2, 4) & ${74.28}_{7.53}$ \\
(1, 4) & ${76.33}_{7.13}$ & (1, 2, 5) & ${77.23}_{6.97}$ \\
(1, 5) & ${76.09}_{7.18}$ & (1, 3, 4) & ${75.94}_{6.91}$ \\
(2, 3) & ${78.91}_{6.88}$ & (1, 3, 5) & ${78.90}_{7.23}$ \\
(2, 4) & ${78.61}_{6.86}$ & (1, 4, 5) & ${73.86}_{7.17}$ \\
(2, 5) & ${76.80}_{6.96}$ & (2, 3, 4) & ${80.15}_{6.59}$ \\
(3, 4) & ${79.63}_{6.76}$ & (2, 3, 5) & ${79.33}_{6.58}$ \\
(3, 5) & ${77.55}_{7.21}$ & (2, 4, 5) & ${76.56}_{7.30}$ \\
(4, 5) & ${77.78}_{7.12}$ & (3, 4, 5) & ${77.52}_{6.92}$ \\ \hline
\end{tabular}%
}
\caption{Results of the focused area model using different topics. We report the mean accuracy and the standard deviation. (1, 2) means focused areas 1 and 2.}
\vspace{-0.3cm}
\label{tab:topic}
\end{table}

We evaluate the focused area model using top-5 focused areas. (\textbf{Picture visualization}) In Figure \ref{fig:topic}, we visualize the top-5 focused areas that have the highest relevance scores. The 1st-rank focused area corresponds to the bottom right area, including the flowing water, sink, and counter. The 2nd focused area covers the 1st area and additionally includes the woman, dish, and window. The 3rd focused area includes the boy and the cookie jar. The 4th focused area is the entire picture. The 5th focused area is the floor area with the flowing water. 

The accuracy and standard deviation of the focused area model are shown in Table \ref{tab:topic} (refer to Table \ref{tab:topic_full} in Appendix for full results). We observe that
i) when the number of samples used for training is small ($\leq 20$), the focused area model performs worse than the baseline model. We consider the focused area model is not effective if the number of sentences to be categorized is small. ii) When the number of samples is large ($> 20$), the focused area model (e.g., (1,2), (1,3), (1,2,3)) outperforms the baseline model, which confirms that the area-based structure of sentences enhances the dementia detection. 
iii) The focused areas should avoid overlapping. For example, using focused areas (1,2) is supposed to achieve higher accuracy than (1,3) due to the higher ranking. However, focused areas (1,2) have a large overlapping region, and categorizing the sentences according to the overlapped focused areas is not effective. iv) Using focused area 4 results in worse performance than the baseline. For example, (1, 4): 76.33\%, (4, 5): 77.78\%, (1, 2, 4): 74.28\%, (1, 3, 4): 75.94\%, and (1, 4, 5): 73.86\%. We consider the worse performance is due to the entire picture as focused area 4; this focused area 4 does not help organize the sentences in a meaningful way. 

\begin{table}[]
\centering
\resizebox{0.45\textwidth}{!}{%
\begin{tabular}{llll}
\hline
ID  & IU     & Focused areas & Frequencies \\ \hline
S1  & boy       & 3,4                  & $5.09\%$            \\
S2  & girl      & 4                       & $4.41\%$            \\
S3  & woman     & 2,4                     & $0.47\%$            \\
S4  & mother    & 2,4                     & $3.65\%$            \\
P1  & kitchen   & 1,2,3,4,5               & $1.29\%$            \\
P2  & exterior  & 2, 4                    & $0.00\%$            \\
O1  & cookie    & 3,4                & $8.42\%$            \\
O2  & jar       & 3,4                 & $5.69\%$            \\
O3  & stool     & 4                       & $5.16\%$            \\
O4  & sink      & 1,2,4                   & $6.12\%$            \\
O5  & plate     & 2,4                     & $1.25\%$            \\
O6  & dishcloth & 1,2,4                   & $0.00\%$            \\
O7  & water     & 1,2,4,5                   & $5.44\%$            \\
O8  & cupboard  & 3,4                   & $0.61\%$            \\
O9  & window    & 2,4                     & $3.04\%$            \\
O10 & cabinet   & 3,4                   & $0.39\%$            \\
O11 & dishes    & 2,4                     & $6.12\%$            \\
O12 & curtains  & 2,4                     & $1.33\%$            \\
O13 & faucet    & 1,2,4                   & $0.36\%$            \\
O14 & floor     & 1,2,4,5                   & $3.01\%$            \\
O15 & counter   & 1,2,4                   & $0.47\%$            \\
O16 & apron     & 1,2,4                   & $0.36\%$            \\ \hline
\end{tabular}%
}
\caption{Information units (IU) and focused areas 1-5. S: Subject; P: Place; O: Object.}
\vspace{-0.3cm}
\label{tab:topic-icu}
\end{table}

We investigate the matching of the focused areas with information units of subjects, places, and objects defined in ~\citep{yancheva-rudzicz-2016-vector}, as shown in Table~\ref{tab:topic-icu}. Focused area 4 covers the whole picture, which includes all information units. Then, we checked the focused areas (1,2,3,5) and found that 20 of all 22 information units are covered by at least one of the focused areas (1,2,3,5). This confirms the consistency between human-defined information units and identified focused areas. In addition, the two information units not covered by the focused areas (1,2,3,5) are "girl" and "stool", which locate in the bottom left area of the picture. On the other hand, we checked that the information units, from high word frequency to low, are cookie ($8.42\%$), sink ($6.12\%$), dishes ($6.12\%$), jar ($5.69\%$), water ($5.44\%$), stool ($5.16\%$), and girl ($4.41\%$). The five top-ranked units are covered by focused areas (1,2,3,5), and thus lower-ranked units "girl" and "stool" in the bottom left area are not covered by the focused areas (1,2,3,5), but are covered by the focused area 7. In addition, the information units with fewer frequencies, e.g., counter (0.47\%), apron (0.36\%), and faucet (0.36\%) are covered by focused areas (1,2,3,5) because their positions are close to the information units with high frequencies. We conclude a fundamental difference between information units and focused areas as follows: for information units extracted from text samples, their high frequencies mean that they were frequently used in participants' description; for the focused areas, their high relevance means some objects inside of the areas have been described by participants, while other objects inside may not be described.

\textbf{Sample visualization.} We show the processed samples of the focused area model in Table~\ref{table:samples}. In this table, we found that sentences in the HC sample are accurately categorized according to focused areas, while some sentences in the AD sample are not. For example, in the AD sample, "upboards underneath the sink" is categorized as focused area 1, while the "upboards underneath the other cupboard" is categorized as focused area 3. Both are supposed to be categorized into focused area 1. We conclude that AD participants may produce more difficult sentences to categorize than HC.

\subsection{Comparison using original evaluation protocol}
{
Compared to the other existing studies, our methods consider integrating information from the cookie theft picture into the model automatically, while most of the other works focus on the information from the speech and the text. In previous works~\cite{balagopalan2020bert, zhu2021exploring, guo2021crossing}, adding a classification layer and fine-tuning achieves 80-83\% accuracy, which is similar performance compared to using SVM (83\%). Guo~\cite{guo2021crossing} used a BERT-based classifier with an external dataset that is not publicly available for fine-tuning. Yuan~\cite{yuan2021pauses} using ERNIE pre-training model  achieved significant best accuracy 89.6\% with 3 pauses features. Li~\cite{li2022grounded} got the best accuracy at 85\% by proposing a method called GPT-D using pre-trained GPT-2 paired with an artificially degraded version of itself to compute the ratio of the perplexities on language from AD and HC participants. As discussed in section 5.3, the original ADReSS evaluation, using fixed training and testing datasets, may result in overestimation. With limited data in this task, fine-tuning may cause the overfitting problem. To compare with existing works, we tested our model using the original ADReSS evaluation protocol and achieved state-of-the-art performance, as shown in Table \ref{tab:compare}. The picture relevance model with (1, 6) achieved the highest accuracy of 89.58\% among our works. The sub-image relevance model with (10, 9) achieved an accuracy of 87.50\%. We conclude that our models achieved higher or equal accuracy than existing works in the original evaluation protocol. }




\begin{table}[]
\centering
\resizebox{0.45\textwidth}{!}{%
\begin{tabular}{llll}
\hline
Model                           & Best accuracy \\ \hline
BERT-based classifier~\cite{guo2021crossing}   & $82.1\%$     \\
Fine-tuned BERT-based classifier (Transfer learning)~\cite{balagopalan2020bert}  & $83.3\%$        \\
ERNIE3p~\cite{yuan2021pauses}  & $\textbf{89.6\%}$     \\
GPT-D~\cite{li2022grounded}  & $\textbf{85}\%$  \\ 
\hline
Picture relevance (1,6)  & $\textbf{89.58\%}$ \\
Sub-image relevance (10,9) & $\textbf{87.50\%}$   \\
Focused area (2, 3, 4)  & $83.33\%$         \\
\hline
\end{tabular}%
}
\caption{{Comparison between existing studies and our models}}
\vspace{-0.3cm}
\label{tab:compare}
\end{table}

\section{Discussion}

\textbf{Limitation of pre-processing.} Our models filter or organize the sentences of samples using their relevance to the picture, sub-image, and focused areas. Alternatively, the relevance scores from the image-text alignment models can be incorporated as parameters into the dementia detection models to maximally preserve the knowledge. 

\textbf{Sentence-level relevance.} The CLIP model has a maximum input length with a limit of 77 tokens. This restriction allows our model to explore only sentence-level relevance. We envision our models would be enhanced with image-text alignment models that could take longer text samples as input. 

\textbf{Focused areas based on text and gaze.} We derived the focused areas using the text description. These focused areas not only include the described objects but also include the non-described objects that are in positions close to the described objects. Without the gaze data, we have no knowledge of whether or not participants have visually focused on these non-described objects. In fact, we envision the visually focused, but non-described objects could play an important role in dementia detection because AD participants may not recall the words to describe the objects. Future work can collect both gaze and text data in the description process to enable the analysis from this aspect.

\section{Conclusion}

In this paper, we explore the picture description dataset for dementia detection by applying an image-text alignment technique. Our models take the cookie theft picture as an input and evaluate the relevance between the picture and the text samples using the knowledge from image-text alignment models. Specifically, we first confirm the picture relevance of HC and AD samples are different. Then, we propose three advanced models where relevance is used to filter or categorize the sentences of samples. We demonstrate that the proposed models (80.63\% picture relevance, 83.44\%
sub-image relevance, 82.49\% focused area) outperform the baseline model (79.91\%). Using picture visualization, we found the left part of the picture is the most dementia-sensitive (83.44\%), and the focused area model using the right part and cookie area as focused areas resulted in the highest accuracy (82.49\%). We confirm the effectiveness of the image-text alignment model in picture description by using sample visualization and correlating human-defined information units and the generated focused areas. Future works include incorporating image-text relevance as parameters of the model instead of filtering and categorizing the samples. Another future work is to develop end-to-end training using the picture as input.


\begin{acks}
This research is funded by the US National Institutes of Health
National Institute on Aging, under grant No. R01AG067416. We thank Professor David Kotz for the feedback about this work.
\end{acks}

\bibliographystyle{ACM-Reference-Format}
\bibliography{reference}

\appendix



\section{Appendix}

\label{sec:appendix}
\begin{table*}[]
\centering
\resizebox{\textwidth}{!}{%
\begin{tabular}{ccccccccc}
\hline
Topic     & 1                 & 5                 & 10                & 20               & 30               & 40               & 50               & 60               \\ \hline
Baseline  & ${60.82}_{12.10}$ & ${72.98}_{8.71}$  & ${75.81}_{7.63}$  & ${77.78}_{7.14}$ & ${78.82}_{6.92}$ & ${79.06}_{6.32}$ & ${79.59}_{6.47}$ & ${79.91}_{7.05}$ \\
(1, 2)    & ${56.87}_{9.72}$  & ${67.34}_{10.19}$ & ${71.51}_{9.48}$  & ${76.72}_{7.57}$ & ${78.61}_{6.70}$ & ${79.94}_{6.75}$ & ${81.15}_{6.33}$ & ${80.83}_{6.62}$ \\
(1, 3)    & ${58.74}_{10.98}$ & ${67.87}_{9.79}$  & ${72.93}_{9.02}$  & ${77.48}_{7.19}$ & ${80.46}_{6.43}$ & ${81.38}_{6.57}$ & ${82.45}_{6.35}$ & ${82.49}_{6.34}$ \\
(1, 4)    & ${57.73}_{10.92}$ & ${65.16}_{10.34}$ & ${69.16}_{9.06}$  & ${73.48}_{7.56}$ & ${74.82}_{7.36}$ & ${75.53}_{7.25}$ & ${75.73}_{7.03}$ & ${76.33}_{7.13}$ \\
(1, 5)    & ${54.36}_{9.30}$  & ${59.47}_{10.31}$ & ${62.71}_{10.57}$ & ${69.96}_{9.35}$ & ${73.62}_{7.67}$ & ${74.97}_{7.41}$ & ${76.02}_{7.09}$ & ${76.09}_{7.18}$ \\
(2, 3)    & ${59.26}_{11.15}$ & ${70.55}_{8.94}$  & ${74.26}_{8.04}$  & ${76.94}_{7.24}$ & ${77.92}_{7.06}$ & ${78.73}_{6.98}$ & ${78.83}_{6.68}$ & ${78.91}_{6.88}$ \\
(2, 4)    & ${57.02}_{10.35}$ & ${64.68}_{10.10}$ & ${69.19}_{9.44}$  & ${74.24}_{7.59}$ & ${75.78}_{7.10}$ & ${77.14}_{7.23}$ & ${78.41}_{6.89}$ & ${78.61}_{6.86}$ \\
(2, 5)    & ${57.23}_{10.34}$ & ${64.02}_{10.47}$ & ${69.61}_{9.28}$  & ${73.12}_{7.57}$ & ${74.36}_{6.90}$ & ${75.91}_{7.05}$ & ${76.54}_{7.17}$ & ${76.80}_{6.96}$ \\
(3, 4)    & ${59.62}_{11.52}$ & ${70.94}_{9.14}$  & ${74.48}_{7.55}$  & ${77.36}_{6.77}$ & ${78.42}_{6.52}$ & ${79.14}_{6.89}$ & ${79.76}_{6.69}$ & ${79.63}_{6.76}$ \\
(3, 5)    & ${57.49}_{10.48}$ & ${67.53}_{9.93}$  & ${71.34}_{8.59}$  & ${74.78}_{7.38}$ & ${76.00}_{7.17}$ & ${76.66}_{7.23}$ & ${77.00}_{7.33}$ & ${77.55}_{7.21}$ \\
(4, 5)    & ${57.54}_{10.42}$ & ${65.31}_{10.24}$ & ${70.87}_{8.85}$  & ${74.97}_{7.72}$ & ${76.47}_{6.73}$ & ${77.04}_{6.97}$ & ${77.65}_{6.77}$ & ${77.78}_{7.12}$ \\
(1, 2, 3) & ${56.40}_{9.69}$  & ${65.14}_{10.45}$ & ${69.59}_{9.78}$  & ${76.01}_{7.69}$ & ${79.12}_{6.75}$ & ${80.63}_{6.53}$ & ${81.19}_{6.43}$ & ${82.24}_{5.93}$ \\
(1, 2, 4) & ${54.04}_{8.49}$  & ${60.53}_{10.05}$ & ${65.03}_{9.93}$  & ${70.14}_{8.14}$ & ${72.72}_{7.65}$ & ${73.51}_{7.30}$ & ${74.49}_{7.19}$ & ${74.28}_{7.53}$ \\
(1, 2, 5) & ${55.44}_{9.06}$  & ${61.02}_{10.06}$ & ${64.76}_{9.60}$  & ${70.09}_{9.00}$ & ${72.98}_{7.53}$ & ${74.93}_{7.74}$ & ${76.24}_{7.00}$ & ${77.23}_{6.97}$ \\
(1, 3, 4) & ${54.92}_{9.18}$  & ${63.99}_{9.81}$  & ${67.36}_{9.28}$  & ${71.46}_{7.48}$ & ${74.01}_{7.52}$ & ${74.74}_{7.12}$ & ${75.02}_{7.13}$ & ${75.94}_{6.91}$ \\
(1, 3, 5) & ${55.13}_{8.95}$ & ${61.16}_{10.61}$ & ${64.58}_{11.22}$ & ${71.33}_{8.64}$ & ${74.63}_{7.25}$ & ${76.06}_{7.32}$ & ${77.95}_{7.18}$ & ${78.90}_{7.23}$ \\
(1, 4, 5) & ${53.98}_{8.88}$  & ${59.06}_{10.08}$ & ${63.59}_{9.93}$  & ${68.73}_{9.21}$ & ${72.03}_{7.64}$ & ${73.24}_{7.03}$ & ${73.87}_{7.22}$ & ${73.86}_{7.17}$ \\
(2, 3, 4) & ${57.03}_{9.86}$  & ${64.26}_{10.24}$ & ${69.62}_{9.16}$  & ${75.49}_{7.55}$ & ${77.52}_{7.31}$ & ${78.81}_{6.92}$ & ${79.70}_{6.42}$ & ${80.15}_{6.59}$ \\
(2, 3, 5) & ${57.47}_{9.64}$  & ${65.75}_{10.30}$ & ${70.37}_{9.07}$  & ${74.98}_{7.62}$ & ${77.07}_{7.04}$ & ${77.42}_{6.80}$ & ${78.56}_{6.31}$ & ${79.33}_{6.58}$ \\
(2, 4, 5) & ${55.58}_{9.58}$  & ${62.69}_{9.83}$  & ${67.56}_{9.47}$  & ${72.43}_{7.89}$ & ${74.36}_{7.69}$ & ${75.23}_{7.27}$ & ${76.18}_{6.95}$ & ${76.56}_{7.30}$ \\
(3, 4, 5) & ${56.76}_{9.86}$  & ${64.64}_{10.70}$ & ${69.97}_{9.17}$  & ${74.38}_{7.63}$ & ${76.42}_{6.85}$ & ${76.98}_{7.34}$ & ${78.16}_{7.04}$ & ${77.52}_{6.92}$ \\ \hline
\end{tabular}%
}
\caption{Results of the focused area model of different shots. We report the mean accuracy and the standard deviation. (1, 2) means using focused areas 1 and 2. }
\label{tab:topic_full}
\end{table*}

\end{document}